\documentclass[11pt,a4paper]{article}
\usepackage[hyperref]{acl2019}
\usepackage{times}
\usepackage{latexsym}
\usepackage{booktabs}
\usepackage{fixltx2e}

\usepackage{url}
\usepackage{graphicx}
\graphicspath{{images/}}
\usepackage{bm}
\usepackage{todonotes}

\usepackage{tikz}
\newcommand*\circled[1]{\tikz[baseline=(char.base)]{
            \node[shape=circle,draw,inner sep=1pt] (char) {#1};}}

\usepackage{float}

\usepackage{subfig}

\aclfinalcopy 

\title{Merge and Label: A novel neural network architecture for nested NER}

\author{Joseph Fisher \\
  Department of Economics \\
  University of Cambridge \\
  \texttt{jhjf2@cam.ac.uk} \\\And
  Andreas Vlachos\\
  Dept. of Computer Science and Technology \\
  University of Cambridge \\
  \texttt{andreas.vlachos@cst.cam.ac.uk} \\}

\date{}

\begin{document}
\maketitle
\begin{abstract}
Named entity recognition (NER) is one of the best studied tasks in natural language processing. However, most approaches are not capable of handling nested structures which are common in many applications.
In this paper we introduce  a novel neural network architecture 
that first merges tokens and/or entities into entities forming nested structures, and then labels each of them independently. 
Unlike previous work, our merge and label approach predicts real-valued instead of discrete segmentation structures, which allow it to combine word and nested entity embeddings while maintaining differentiability. 
 We evaluate our approach using the ACE 2005 Corpus, where it achieves state-of-the-art F1 of 74.6, further improved with contextual embeddings (BERT) to 82.4, an overall improvement of close to 8 F1 points over previous approaches trained on the same data. Additionally we compare it against BiLSTM-CRFs, the dominant approach for flat NER structures, demonstrating that its ability to predict nested structures does not impact performance in simpler cases.\footnote{Code available at \url{https://github.com/fishjh2/merge_label}}
 
\end{abstract}

\section{Introduction}

\begin{figure*}[!th]
	\includegraphics[width=1.0\textwidth]{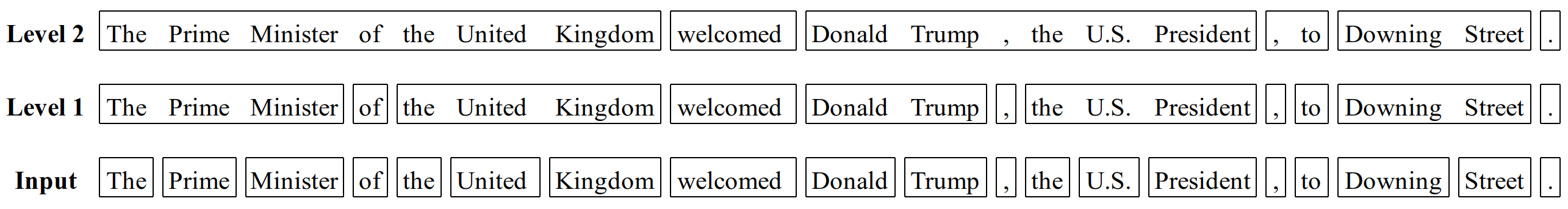}
	\caption{Trained model's representation of nested entities, after thresholding the merge values, $M$ (see section 2.1). Note that the merging of ``, to'' is a mistake by the model.}
	\label{fig:rep}
\end{figure*}

The task of nested named entity recognition (NER)
focuses on recognizing and classifying entities that can be nested within each other, such as  ``United Kingdom'' and ``The Prime Minister of the United Kingdom'' in Figure~\ref{fig:rep}.
Such entity structures, while very commonly occurring, cannot be handled by the predominant variant of NER models \citep{mccallum2003early,lample2016neural}, which can only tag non-overlapping  entities.

A number of approaches have been proposed for nested NER.
\citet{Roth} introduced a hypergraph representation which can represent 
overlapping mentions, which was further improved by \citet{Gaps}, by assigning tags between each pair of consecutive words, preventing the model from learning spurious structures (overlapping entity structures which are gramatically impossible). More recently, \citet{Revisited} built on this approach, adapting an LSTM \cite{Hochreiter:1997:LSM:1246443.1246450} to learn the hypergraph directly, and \citet{Segmental} introduced a segmental hypergraph approach, which is able to incorporate a larger number of span based features, by encoding each span with an LSTM.

Our approach decomposes nested NER into two stages. First tokens are merged into entities (Level 1 in Figure~\ref{fig:rep}), which are merged with other tokens or entities in higher levels. These merges are encoded as real-valued decisions, which enables a parameterized combination of word embeddings into entity embeddings at different levels. These entity embeddings are used to label the entities identified. The model itself consists of feedforward neural network layers and is fully differentiable, thus it is straightforward to train with backpropagation.

Unlike methods such as \citet{Revisited}, it does not predict entity segmentation at each layer as discrete 0-1 labels, thus allowing the model to flexibly aggregate information across layers.
Furthermore inference is greedy, without attempting to score all possible entity spans as in \citet{Segmental}, which results in faster decoding (decoding requires simply a single forward pass of the network). 

To test our approach on nested NER, we evaluate it on the ACE 2005 corpus (LDC2006T06) where it achieves a state-of-the-art F1 score of 74.6. This is further improved with contextual embeddings \citep{bert} to 82.4, an overall improvement of close to 8 F1 points against the previous best approach trained on the same data, \cite{Segmental}. 
Our approach is also 60 times faster than its closest competitor. Additionally, we compare it against BiLSTM-CRFs\cite{BILSTM_CRF}, the dominant flat NER paradigm, on Ontonotes (LDC2013T19) and demonstrate that its ability to predict nested structures does not impact performance in flat NER tasks as it achieves comparable results to the state of the art on this dataset.


\section{Network Architecture} \label{architecture}

\subsection{Overview}

\begin{figure}[h]
	\includegraphics[width=0.45\textwidth]{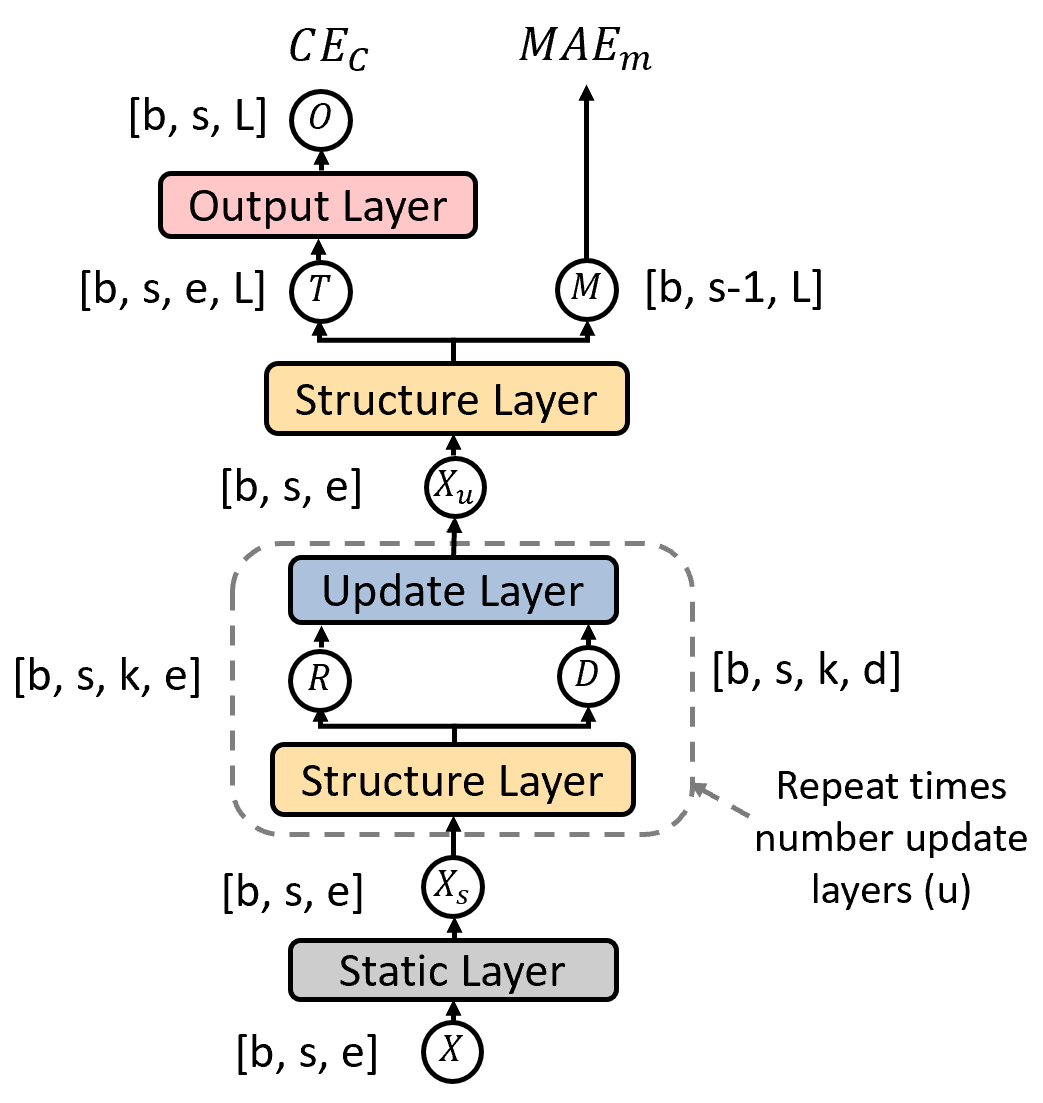}
	\caption{Model architecture overview}
	\label{fig:sentshape}
\end{figure}

The model decomposes nested NER into two stages. Firstly, it identifies the boundaries of the named entities at all levels of nesting; the tensor M in Figure \ref{fig:sentshape}, which is composed of real values between 0 and 1 (these real values are used to infer discrete split/merge decisions at test time, giving the nested structure of entities shown in Figure~\ref{fig:rep}). We refer to this as predicting the ``structure" of the NER output for the sentence. 
Secondly, given this structure, it produces embeddings for each entity, by combining the embeddings of smaller entities/tokens from previous levels (i.e. there will be an embedding for each rectangle in Figure \ref{fig:rep}). These entity embeddings are used to label the entities identified.

An overview of the architecture used to predict the structure and labels is shown in Figure \ref{fig:sentshape}. The dimensions of each tensor are shown in square brackets in the figure. The input tensor, $X$, holds the word embeddings of dimension
$e$, for every word in the input of sequence length, $s$. The first dimension, $b$, is the batch size. The \textbf{Static Layer} updates the token embeddings using contextual information, giving tensor $X_s$ of the same dimension, $[b, s, e]$.

Next, for $u$ repetitions, we go through a series of building the 
structure using the \textbf{Structure Layer}, and then use this structure to continue updating the individual token embeddings using the \textbf{Update Layer}, giving an output $X_u$. 

The updated token embeddings $X_u$ are passed through the Structure Layer one last time, to give the final entity embeddings, $T$ and structure, $M$. 
%
A feedforward \textbf{Output Layer} then gives the predictions of the label of each entity. 

The structure is represented by the tensor $M$, of dimensions $[b, s-1, L]$. $M$ holds, for every pair of adjacent words ($s-1$ given input length $s$) and every output level ($L$ levels), a value between 0 and 1. A value close to 0 denotes that the two (adjacent) tokens/entities from the previous level are likely to be merged on this level to form an entity; nested entities emerge when entities from lower levels are used. Note that for each individual application of the Structure Layer, we are building multiple levels (L) of nested entities. That is, within each Structure Layer there is a loop of length L. 
By building the structure before the Update Layer, the updates to the token embeddings can utilize information about which entities each token is in, as well as neighbouring entities, as opposed to just using information about neighbouring tokens. 


\subsection{Preliminaries}

Before analysing each of the main layers of the network, we introduce two building blocks, which are used multiple times throughout the architecture. The first one is the \textbf{Unfold operators}. Given that we process whole news articles in one batch (often giving a sequence\_length (s) of 500 or greater) we do not allow each token in the sequence to consider 
every other token. Instead, we define a kernel of size k around each token, similar to convolutional neural networks \citep{kimconvolutional}, allowing it to 
consider the k/2 prior tokens and the k/2 following tokens. 

\begin{figure}[h]
	\includegraphics[width=0.5\textwidth]{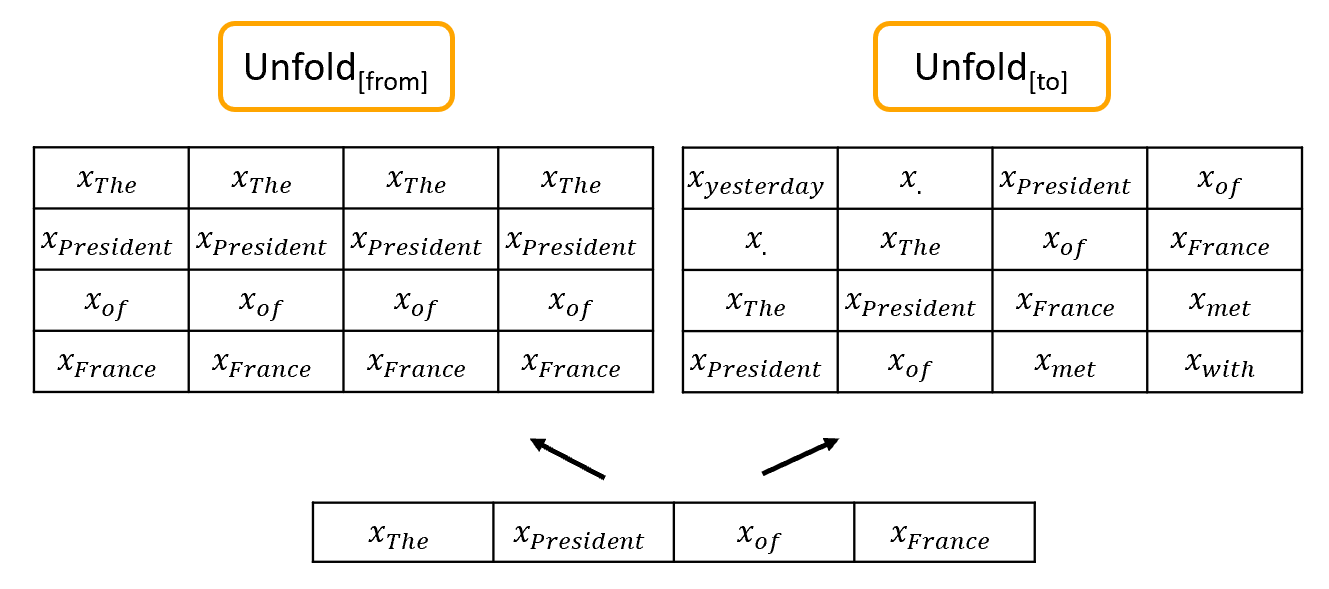}
	\caption{Unfold Operators for the  passage ``... yesterday. The President of France met with ...". Each row in the matrices corresponds to the words ``The'', ``President'', ''of'' and ``France'' (top to bottom).}
	\label{fig:unfold}
\end{figure}

The unfold operators create 
kernels transforming tensors holding the word embeddings of shape [b, s, e] to shape [b, s, k, e]. unfold\textsubscript{[from]} simply tiles the embedding $x$ of each token k times, and unfold\textsubscript{[to]} generates the k/2 token embeddings either side, as shown in Figure~\ref{fig:unfold}, for a kernel size $k$ of 4. The first row of the unfold\textsubscript{[to]} tensor holds the two tokens before and the two tokens after the word ``The", the second row the two before and after ``President" etc. As we process whole articles, the unfold operators allow tokens to 
consider tokens from previous/following sentences. 

\begin{figure}[h]
    \begin{center}
	\includegraphics[width=0.3\textwidth]{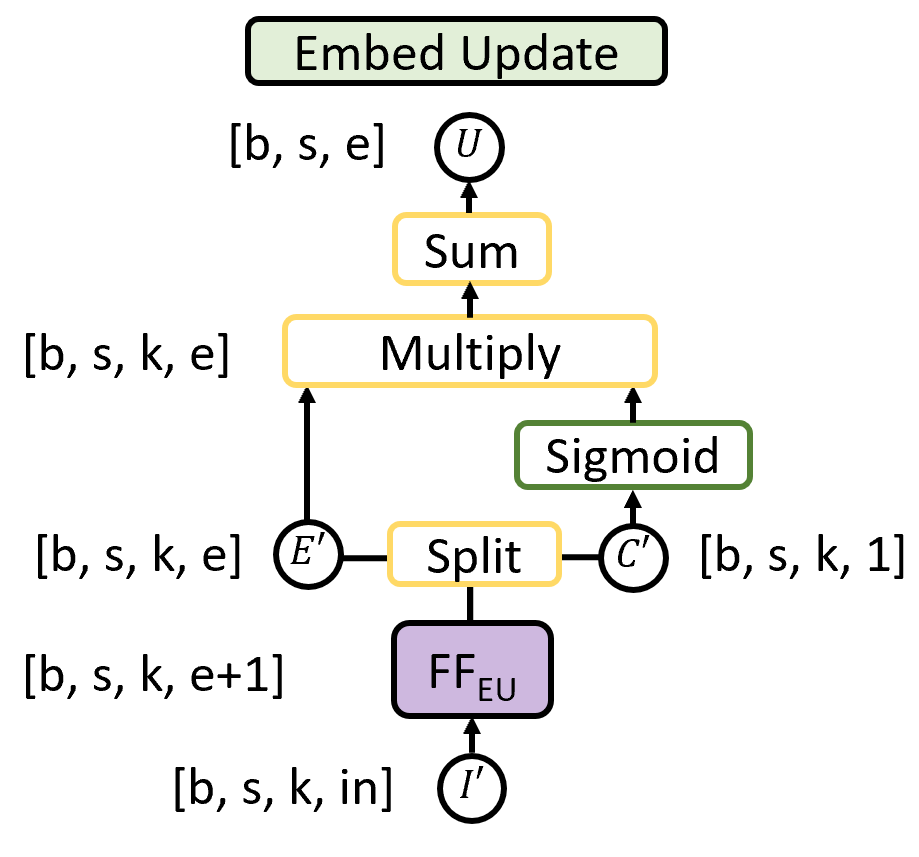}
	\caption{Embed Update layer}
	\label{fig:embed_update}
	\end{center}
\end{figure}

The second building block is the \textbf{Embed Update layer}, shown in Figure~\ref{fig:embed_update}. This layer is used to update embeddings 
within the model, and as such, can be thought of as equivalent in function to the residual update mechanism in Transformer \cite{transformer}. It is used in each of the Static Layer, Update Layer and Structure Layer from the main network architecture in Figure \ref{fig:sentshape}.

It takes an input $I'$ of size $[b, s, k, in]$, formed using the unfold ops described above, where the last dimension $in$ varies depending on the point in the architecture at which the layer is used.
It passes this input through the feedforward NN $FF_{EU}$, 
giving an output of dimension $[b, s, k, e+1]$ (the network broadcasts over the last three dimensions of the input tensor). The output is split into two. Firstly, a tensor $E'$ of shape $[b, s, k, e]$, which holds, for each word in the sequence, k predictions of an updated word vector based on the k/2 words either side. Secondly, a weighting tensor $C'$ of shape $[b, s, k, 1]$, which is scaled between 0 and 1 using the sigmoid function, and denotes how ``confident" each of the $k$ predictions is about its update to the word embedding. This works similar to an attention mechanism, allowing each token to focus on updates from the most relevant neighbouring tokens.\footnote{The difference being that the weightings are generated using a sigmoid rather than a softmax layer, allowing the attention values to be close to one for multiple tokens.} The output, $U$ is then a weighted average of $E'$:

\[U = sum_2(sigmoid(C') * E') \]

where $sum_2$ denotes summing across the second dimension of size $k$. $U$ therefore has dimensions $[b, s, e]$ and contains the updated embedding for each word.

During training we initialize the weights of the network using the identity function. As a result, the default behaviour of  $FF_{EU}$ 
prior to training is to pass on the word embedding unchanged, which is then updated during via backpropagation. An example of the effect of the identity initialization is provided in the supplementary materials.

\subsection{Static Layer}

The static layer is a simple preliminary layer to update the embeddings for each word based on contextual information, and as such, is very similar to a Transformer \cite{transformer} layer. Following the unfold ops, a positional encoding $P$ of dimension $e$ (we use a learned encoding) is added, giving tensor $I_s$:

\[I_s = concat(Unfold_{[from]}(X), Unfold_{[to]}(X) + P)\]

$I_s$ is then passed through the Embed Update layer. In our experiments, we use a single static layer. There is no merging of embeddings into entities in the static layer.

\begin{figure}[h]
    \begin{center}
	\includegraphics[width=0.40\textwidth]{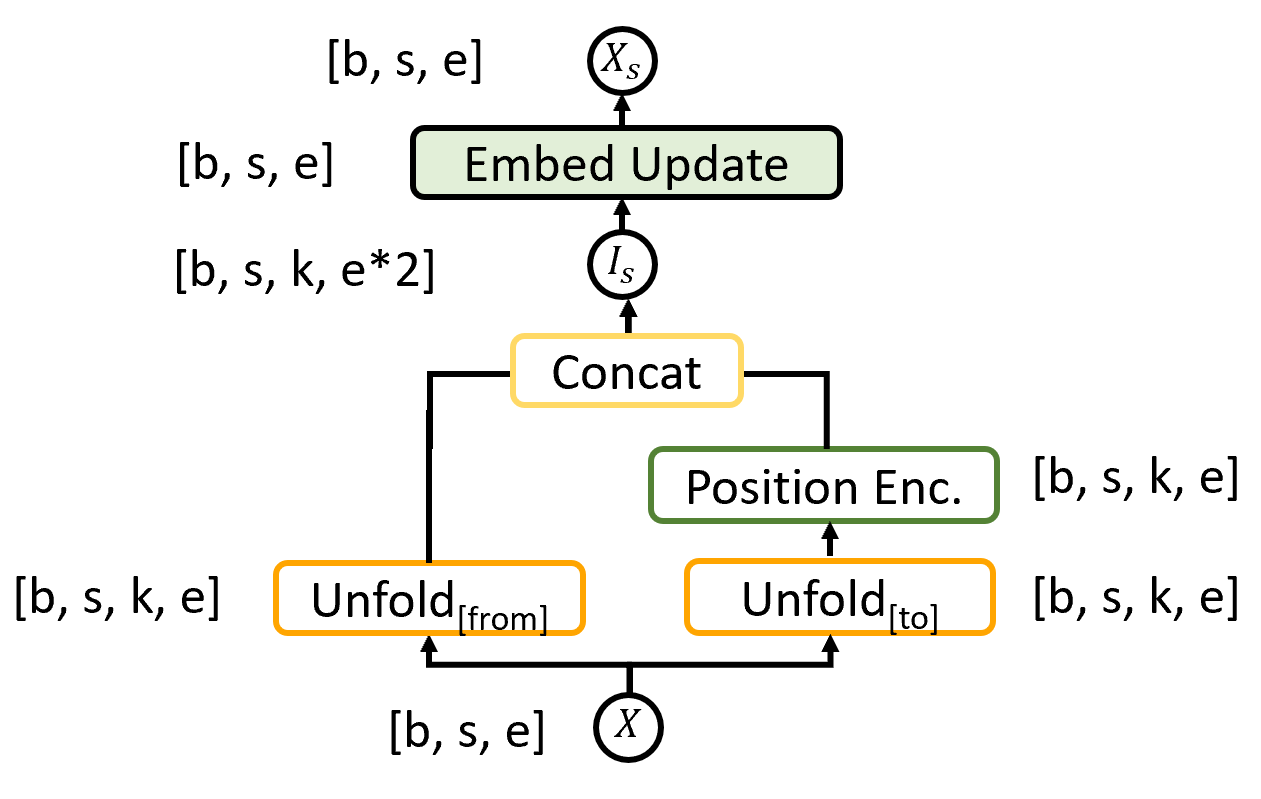}
	\caption{Static Layer}
	\label{fig:static_layer}
	\end{center}
\end{figure}

\subsection{Structure Layer}

The Structure Layer is responsible for three tasks. Firstly, deciding which token embeddings should be merged at each level, expressed as real values between 0 and 1, and denoted $M$. 
Secondly, given these merge values $M$, deciding how the separate token embeddings should be combined in order to give the embeddings for each entity, $T$. Finally, for each token and entity, providing directional vectors  $D$ to the $k/2$ tokens either side, which are used to update each token embedding in the Update Layer based on its context. Intuitively, the directional vectors $D$ can be thought of as encoding relations between entities - such as the relation between an organization and its leader, or that between a country and its capital city (see Section \ref{dir_disc} for an analysis of these relation embeddings).

 \begin{figure}[h]
	\includegraphics[width=0.49\textwidth]{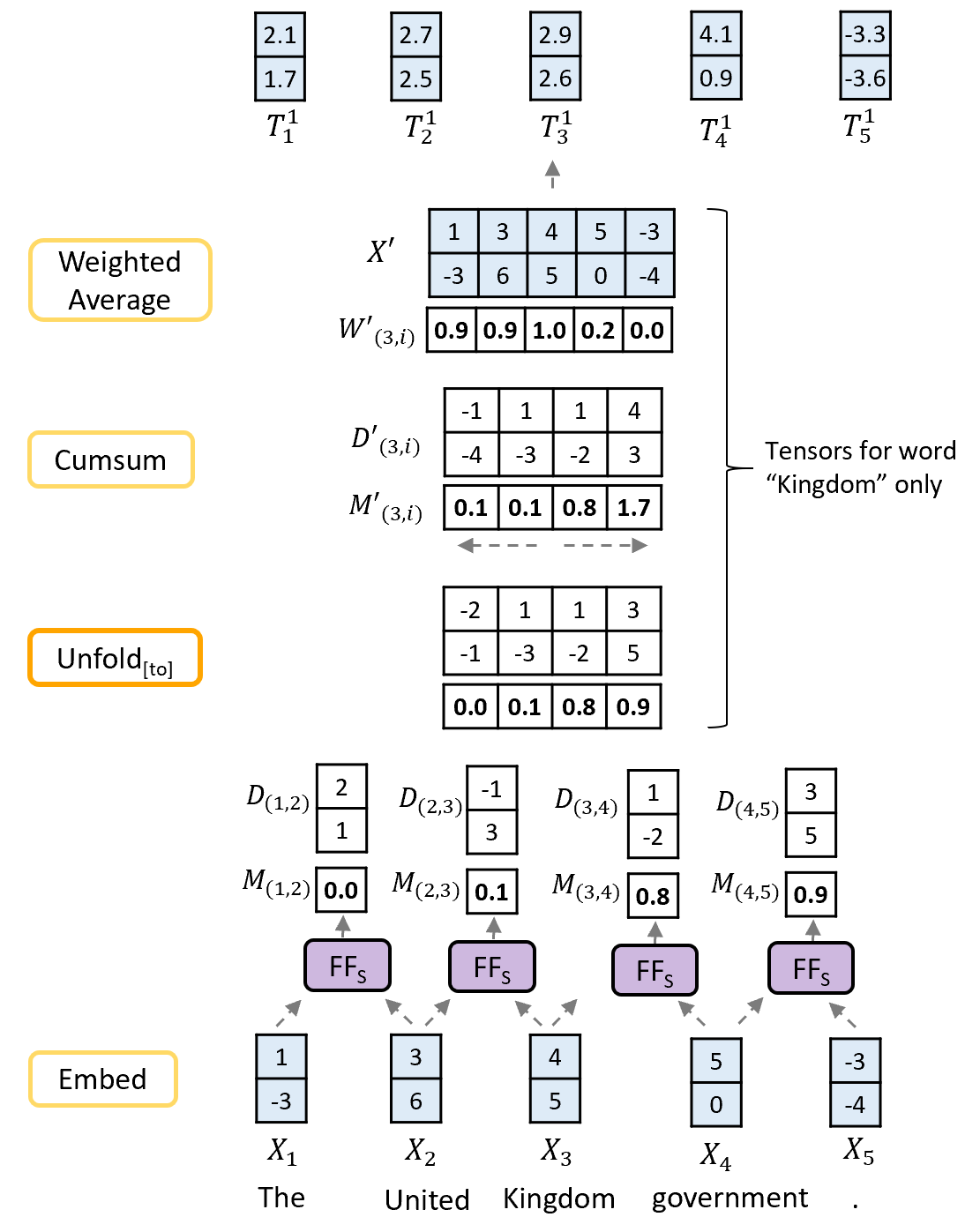}
	\caption{Calculation of merging weight, directions and entities in Structure Layer}
	\label{fig:sees}
\end{figure}

Figure \ref{fig:sees} shows a minimal example of the calculation of $D$, $M$ and $T$, with word embedding and directional vector dimensions $e=d=2$, and kernel size, $k = 4$. We pass the embeddings ($X$) of each pair of adjacent words through a feedforward NN $FF_S$ 
to give directions $D$ [b, s-1, d] and merge values $M$ [b, s-1, 1] between each pair. If $FF_S$ 
predicts $M_{(1,2)}$ to be close to 0, this indicates that tokens 1 and 2 are part of the same entity on this level.
The unfold\textsubscript{[to]} op gives, for each word (we show only the unfolded tensors for the word ``Kingdom" in Figure \ref{fig:sees} for simplicity), $D$ and $M$ for pairs of words up to k/2 either side.

By taking both the left and right cumulative sum (cumsum) of the resulting two tensors from the center out (see grey dashed arrows in Figure \ref{fig:sees} for direction of the two cumsum ops), we get directional vectors and merge values from the word ``Kingdom" to the words before and after it in the phrase, $D'_{3,i}$ and $M'_{3, i}$ for $i = (1,2,4,5)$. Note that we take the inverse of vectors $D_{(1, 2)}$ and $D_{(2,3)}$ prior to the cumsum, as we are interested in the directions \textbf{from} the token ``Kingdom" backwards to the tokens ``United" and ``The". The values $M'_{3, i}$ are converted to weights $W'$ of dimension [b, s, k, 1] using the formula $W' = max(0, 1 - M')$\footnote{We use the notation $D'$ to denote the unfolded version of tensor $D$, i.e.\ $D' = Unfold_{[to]}(D)$}, with the max operation ensuring the model puts a weight of zero on tokens in separate entities (see the reduction of the value of 1.7 in $M'$ in Figure \ref{fig:sees} to a weighting of 0.0). The weights are normalized to sum to 1, and multiplied with the unfolded token embeddings $X'$ to give the entity embeddings $T$, of dimension [b, s, e] 

\[T = \frac{W'}{sum_{2}(W')} * X'\]
Consequently, the embeddings at the end of level 1 for the words ``The", ``United" and ``Kingdom" ($T^1_1$, $T^1_2$ and $T^1_3$ respectively) are all now close to equal,  and all have been formed from a weighted average of the three separate token embeddings. If $M_{(1,2)}$ and $M_{(2,3)}$ were precisely zero, and $M_{(3,4})$ was precisely 1.0, then all three would be identical. In addition, on higher levels, the directions from other words to each of these three tokens will also be identical. In other words, the use of ``directions"\footnote{We use the term ``directions" as we inverse the vectors to get the reverse direction, and cumsum them to get directions between tokens multiple steps away.} allows the network to represent entities as a single embedding in a fully differentiable fashion, whilst keeping the sequence length constant.

Figure \ref{fig:sees} shows just a single level from within the Structure Layer. The embeddings $T$ are then passed onto the next level, allowing progressively larger entities to be formed by combining smaller entities from the previous levels. 

\begin{figure}[h]
	\includegraphics[width=0.5\textwidth]{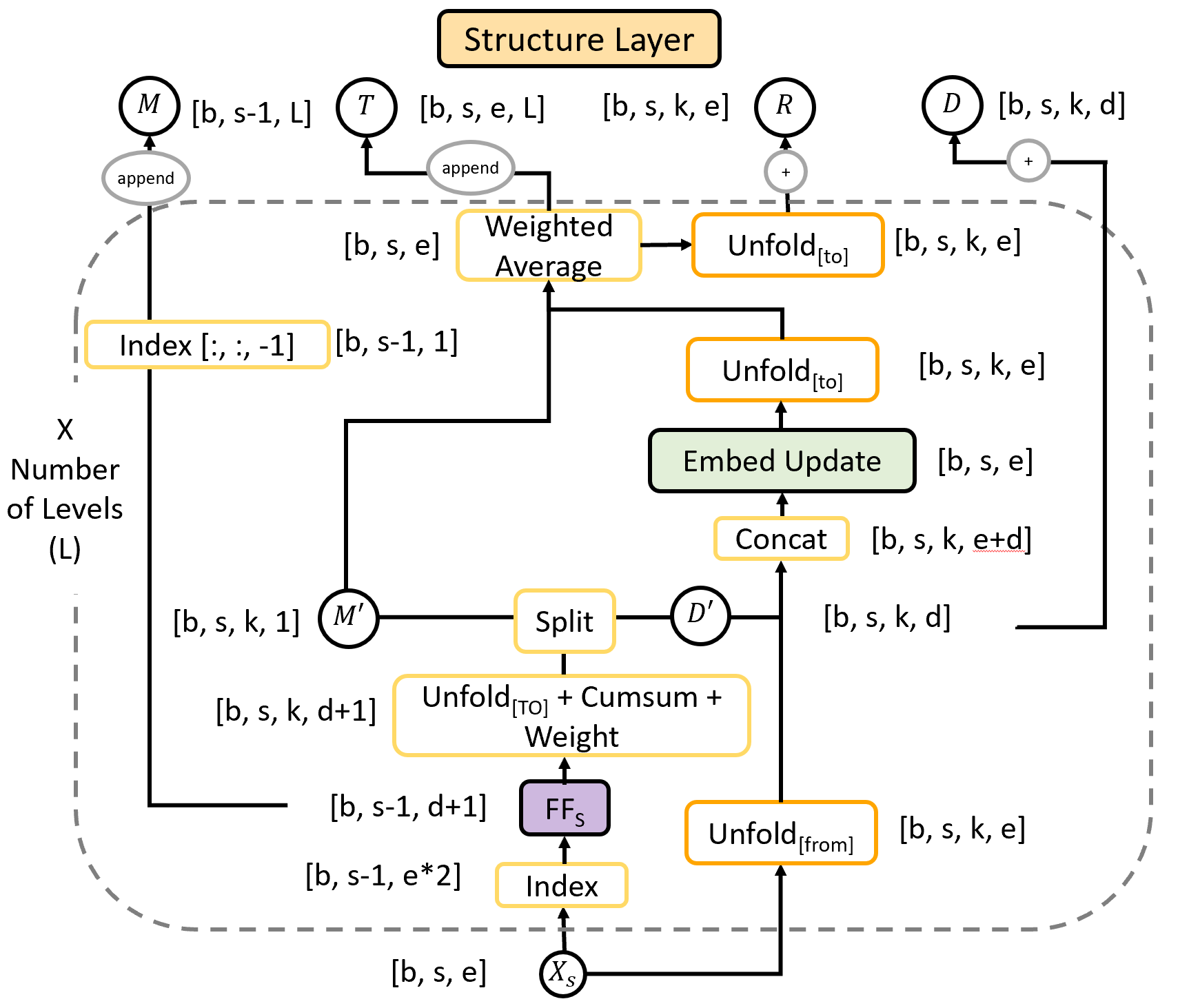}
	\caption{Structure Layer}
	\label{fig:levels_layer}
\end{figure}

The full architecture of the Structure Layer is shown in Figure \ref{fig:levels_layer}. The main difference to Figure \ref{fig:sees} is the additional use of Embed Update Layer, to decide how individual token/entity embeddings are combined together into a single entity. 
The reason for this is that if we are joining the words ``The", ``United" and ``Kingdom" into a single entity, it makes sense that the joint vector should be based largely on the embeddings of ``United" and ``Kingdom", as ``The" should add little information. 
The embeddings are unfolded (using the unfold\textsubscript{[from]} op) to shape $[b, s, k, e]$ and concatenated with the directions between words, $D'$, to give the tensor of shape $[b, s, k, e+d]$. This is passed through the Embed Update layer, giving, for each word, a weighted and updated embedding, ready to be combined into a single entity (for unimportant words like ``The", this embedding will have been reduced to close to zero). We use this tensor in place of tensor $X$ in Figure \ref{fig:sees}, and multiply with the weights $W'$ to give the new entity embeddings, $T$.

There are four separate outputs from the Structure Layer. The first, denoted by \circled{$T$}, is the entity embeddings from each of the levels concatenated together, giving a tensor of size [b, s, e, L]. The second output, \circled{$R$}, is a weighted average of the embeddings from different layers, of shape $[b, s, k, e]$. This will be used in the place of the unfold\textsubscript{[to]} tensor described above as an input the the Update Layer. It holds, for each token in the sequence, embeddings of entities up to $k/2$ tokens either side. 
The third output, \circled{$D$}, will also be used by the Update Layer. It holds the directions of each token/entity to the $k/2$ tokens/entities either side. It is formed using the cumsum op, as shown in Figure \ref{fig:sees}. Finally, the fourth output, \circled{$M$}, stores the merge values for every level. It is used in the loss function, to directly incentivize the correct merge decisions at the correct levels. 

\subsection{Update Layer}

The Update Layer is responsible for updating the \textbf{individual} word vectors, using the contextual information derived from outputs \circled{R} and \circled{D} of the Structure Layer. It concatenates the two outputs together, along with the output of the unfold\textsubscript{[from]} op, $X_s'$, and with an article theme embedding $A$ tensor, giving tensor $Z$ of dimension [b, s, k, (e*2 + d + a)]. The article theme embedding is formed by passing every word in the article through a feedforward NN, and taking a weighted average of the outputs, giving a tensor of dimension $[b, a]$. This is then tiled\footnote{Tiling refers to simply repeating the tensor across \textbf{both} the sequence length $s$ \textbf{and} kernel size $k$ dimensions} to dimension $[b, s, k, a]$, giving tensor $A$. $A$ allows the network to adjust its contextual understanding of each token based on whether the article is on finance, sports, etc. $Z$ is then passed through an Embed Update layer, giving an output $X_u$ of shape $[b, s, e]$. 

\[X_u = Embed\_Update(concat(X_s', R, D, A))\]

We therefore update each word vector using four pieces of information. The original word embedding, a direction to a different token/entity, the embedding of that different token/entity, and the article theme. 

\begin{figure}[h]
	\begin{center}
	\includegraphics[width=0.4\textwidth]{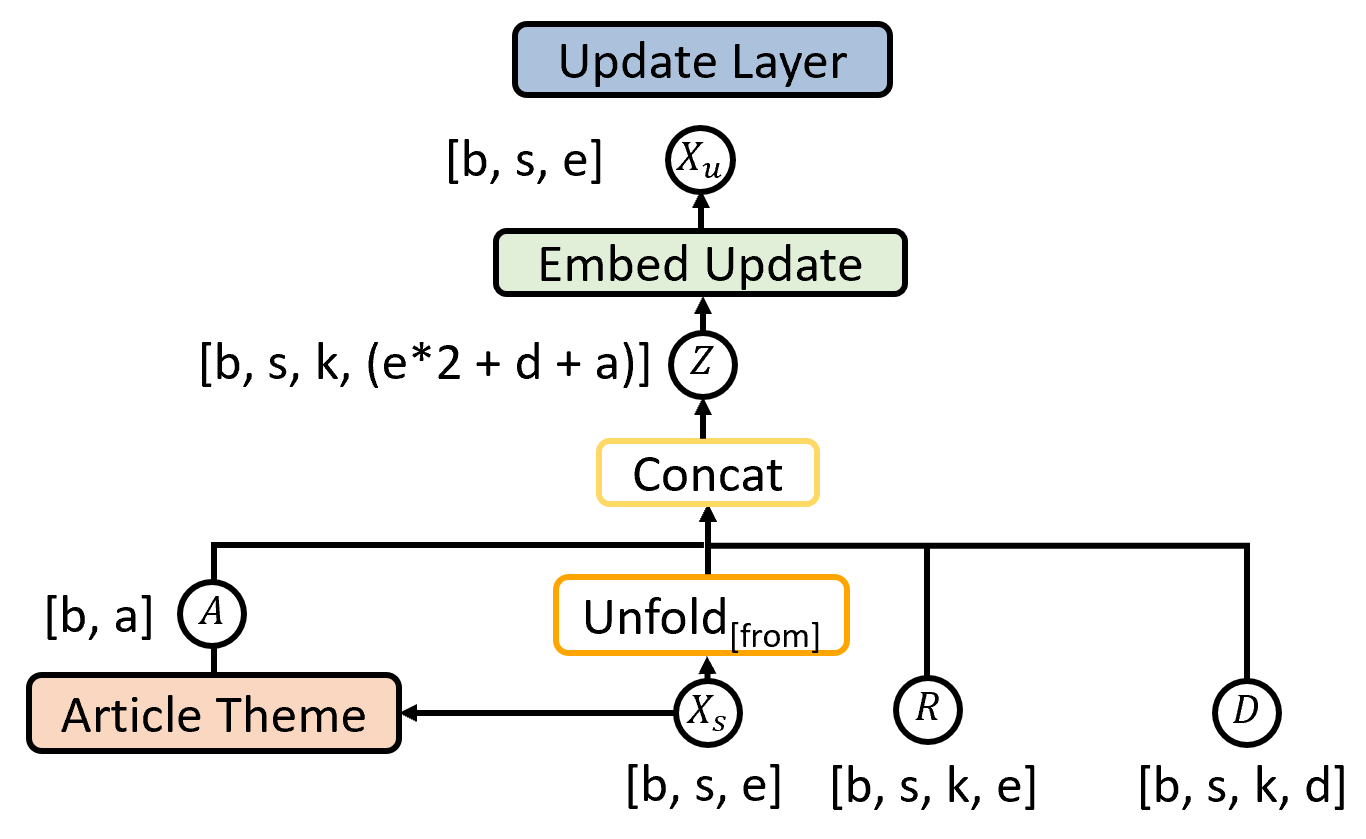}
	\caption{Update Layer}
	\label{fig:update_layer}
	\end{center}
\end{figure}

The use of directional vectors $D$ in the Update Layer can be thought of as an alternative to the positional encodings in Transformer \cite{transformer}. That is, instead of updating each token embedding using neighbouring tokens embeddings with a positional encoding, we update using neighbouring token embeddings, and the \textbf{directions} to those tokens. 

\section{Implementation Details}

\subsection{Data Preprocessing}

\subsubsection{ACE 2005}

ACE 2005 is a corpus of around 180K tokens, with 7 distinct entity labels. The corpus labels include nested entities, allowing us to compare our model to the nested NER literature. The dataset is not pre-tokenized, so we carry out sentence and word tokenization using NLTK.

\subsubsection{OntoNotes}

\begin{figure}[h]
	\includegraphics[width=0.46\textwidth]{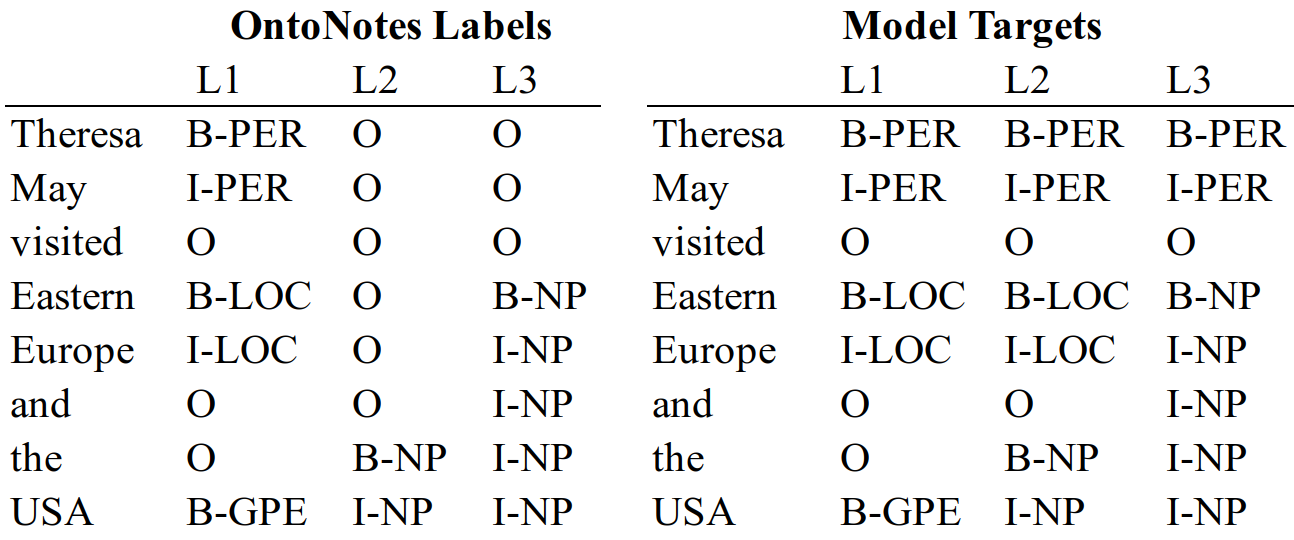}
	\caption{OntoNotes Labelling}
	\label{fig:labels}
\end{figure}

OntoNotes v5.0 is the largest corpus available for NER, comprised of around 1.3M tokens, and 19 different entity labels. Although the labelling of the entities is not nested in OntoNotes, the corpus also includes labels for all noun phrases, which we train the network to identify concurrently. For training, we copy entities which are not contained within a larger nested entity onto higher levels, as shown in Figure \ref{fig:labels}.

\subsubsection{Labelling}

For both datasets, during training, we replace all ``B-" labels with their corresponding ``I-" label. At evaluation, all predictions which are the first word in a merged entity have the ``B-" added back on. 
As the trained model's merging weights, $M$, can take any value between 0 and 1, we have to set a cutoff at eval time when deciding which words are in the same entity. We perform a grid search over cutoff values using the dev set, with a value of 0.75 proving optimal.

\subsection{Loss function}

The model is trained to predict the correct merge decisions, held in the tensor $M$ of dimension [b, s-1, L] and the correct class labels given these decisions, $C$. The merge decisions are trained directly using the mean absolute error (MAE):


\[MAE_M =  \frac{sum(| M - \hat{M} |)}{(b * s * L)} \]

This is then weighted by a scalar $w_M$, and added to the usual Cross Entropy (CE) loss from the predictions of the classes, ${CE}_C$, giving a final loss function of the form:

\[Loss = (w_{M} * MAE_{M})  + {CE}_C\]

In experiments we set the weight on the merge loss, $w_{M}$ to 0.5.

\subsection{Evaluation}

Following previous literature, for both the ACE and OntoNotes datasets, we use a strict F1 measure, where an entity is only considered correct if both the label and the span are correct. 

\subsubsection{ACE 2005}

For the ACE corpus, the default metric in the literature \cite{Wang,Meizhi,Segmental} does not include sequential ordering of nested entities (as many architectures do not have a concept of ordered nested outputs). As a result, an entity is considered correct if it is present in the target labels, regardless of which layer the model predicts it on. 

\subsubsection{OntoNotes}

NER models evaluated on OntoNotes are trained to label the 19 entities, and not noun phrases (NP). To provide as fair as possible a comparison, we consequently flatten all labelled entities into a single column. As 96.5\% of labelled entities in OntoNotes do not contain a NP nested inside, this applies to only 3.5\% of the dataset. 

\begin{figure}[h]
    \begin{center}
	\includegraphics[width=0.4\textwidth]{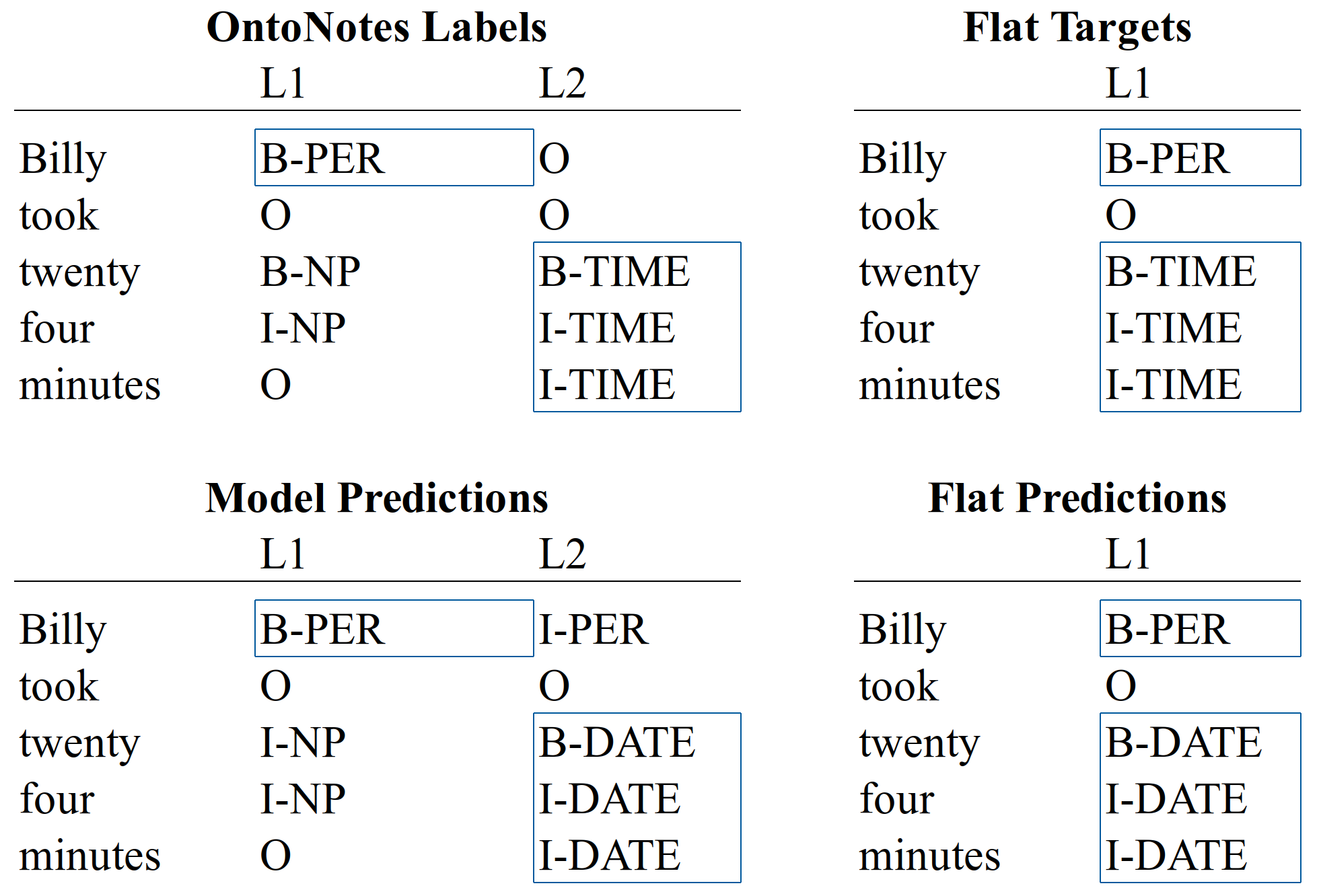}
	\end{center}
	\caption{OntoNotes Targets}
	\label{fig:flat_labels}
\end{figure}

The method used to flatten the targets is shown in Figure \ref{fig:flat_labels}. The OntoNotes labels include a named entity (TIME), in the second column, with the NP ``twenty-four" minutes nested inside. Consequently, we take the model's prediction from the second column as our prediction for this entity. This provides a fair comparison to existing NER models, as all entities are included, and if anything, disadvantages our model, as it not only has to predict the correct entity, but do so on the correct level. That said, the NP labels provide additional information during training, which may give our model an advantage over flat NER models, which do not have access to these labels. 

\subsection{Training and HyperParameters}

We performed a small amount of hyperparameter tuning across dropout, learning rate, distance embedding size $d$, and number of update layers $u$. We set dropout at 0.1, the learning rate to 0.0005, $d$ to 200, and $u$ to 3. For full hyperparameter details see the supplementary materials. The number of levels, $L$, is set to 3, with a kernel size $k$ of 10 on the first level, 20 on the second, and 30 on the third (we increase the kernel size gradually for computational efficiency as first level entities are extremely unlikely to be composed of more than 10 tokens, whereas higher level nested entities may be larger). Training took around 10 hours for OntoNotes, and around 6 hours for ACE 2005, on an Nvidia 1080 Ti. 

For experiments without language model (LM) embeddings, we used pretrained Glove embeddings \cite{glove} of dimension 300. Following \cite{Strubell}, we added a ``CAP features" embedding of dimension 20, denoting if each word started with a capital letter, was all capital letters, or had no capital letters. For the experiments with LM embeddings, we used the implementations of the BERT \cite{bert} and ELMO \cite{ELMO} models from the Flair \cite{Flair} project\footnote{https://github.com/zalandoresearch/flair/}. We do not finetune the BERT and ELMO models, but take their embeddings as given.
  
\section{Results}

\subsection{ACE 2005}

On the ACE 2005 corpus, we begin our analysis of our model's performance by comparing to models which do not use the POS tags as additional features, and which use non-contextual word embeddings. These are shown in the top section of Table \ref{table:ace}. The previous state-of-the-art F1 of 72.2 was set by \citet{Meizhi}, using a series of stacked BiLSTM layers, with CRF decoders on top of each of them. Our model improves this result with an F1 of 74.6 (avg. over 5 runs with std. dev. of 0.4). This also brings the performance into line with 
\citet{Wang} and \citet{Segmental}, 
which concatenate embeddings of POS tags with word embeddings as an additional input feature.
%

\begin{table}[h]
    \small
    \begin{center}
	\setlength{\tabcolsep}{2pt}
	\begin{tabular}{lccc}
		\textbf{Model} & \textbf{Pr.} & \textbf{Rec.} & \textbf{F1} \\
		\midrule
		Multigraph + MS \cite{Gaps} & 69.1 & 58.1 & 63.1 \\
		RNN + hyp \cite{Revisited} & 70.6 & 70.4 & 70.5 \\
		BiLSTM-CRF stacked \cite{Meizhi} & 74.2 & 70.3 & 72.2 \\
		LSTM + forest [POS] \cite{Wang} & 74.5 & 71.5 & 73.0 \\
		Segm. hyp [POS] \cite{Segmental} & 76.8 & 72.3 & 74.5 \\
		Merge and Label & 75.1 & 74.1 & \textbf{74.6} \\
		& & & \\
		LM embeddings & & & \\
		\midrule
		Merge and Label [ELMO] & 79.7 & 78.0 & 78.9\\
		Merge and Label [BERT] & 82.7 & 82.1 & \textbf{82.4}\\
	    & & & \\
	    LM + OntoNotes & & & \\
	    \midrule
	    DyGIE \cite{general_frame} & &  & \textbf{82.9} \\
	\end{tabular}
	\caption{ACE 2005}
	\label{table:ace}
	\end{center}
\end{table}

Given the recent success on many tasks using contextual word embeddings, we also evaluate performance using the output of pre-trained BERT \cite{bert} and ELMO \cite{ELMO} models as input embeddings. This leads to a significant jump in performance to 78.9 with ELMO, and 82.4 with BERT (both avg. over 5 runs with 0.4 and 0.3 std. dev. respectively), an overall increase of 8 F1 points from the previous state-of-the-art. Finally, we report the concurrently published result of \citet{general_frame}, in which they use ELMO embeddings, and additional labelled data (used to train the coreference part of their model and the entity boundaries) from the larger OntoNotes dataset.

A secondary advantage of our architecture relative to those models which require construction of a hypergraph or CRF layer is its decoding speed, as decoding requires only a single forward pass of the network. As such it achieves a speed of 9468 words per second (w/s) on an Nvidia 1080 Ti GPU, relative to a reported
speed of 157 w/s for the closest competitor model of \citet{Segmental}, a sixty fold advantage.

\subsection{OntoNotes}

As mentioned previously, given the caveats that our model is trained to label all NPs as well as entities, and must also predict the correct layer of an entity, the results in Table \ref{table:ontonotes} should be seen as indicative comparisons only. Using non-contextual embeddings, our model achieves a test F1 of 87.59. To our knowledge, this is the first time that a nested NER architecture has performed comparably to BiLSTM-CRFs \cite{BILSTM_CRF} (which have dominated the named entity literature for the last few years) on a flat NER task.

Given the larger size of the OntoNotes dataset, we report results from a single iteration, as opposed to the average of 5 runs as in the case of ACE05.

\begin{table}[h]
\small
\setlength{\tabcolsep}{3pt}
    \begin{center}
	\begin{tabular}{lc}
		\textbf{Model} & \textbf{F1} \\[5pt]
		\midrule
		BiLSTM-CRF \cite{Chiu_Nichols} & 86.28 \\
		ID-CNN \cite{Strubell} & 86.84 \\
		BiLSTM-CRF \cite{Strubell} & 86.99 \\
		Merge and Label & \textbf{87.59} \\
		 & \\
		LM embeddings or extra data & \\
		\midrule
		BiLSTM-CRF lex \cite{Ghaddar} & 87.95 \\
		BiLSTM-CRF with CVT \cite{CVT} & 88.81 \\
		Merge and Label [BERT] & 89.20 \\
		BiLSTM-CRF Flair \cite{Flair}  & \textbf{89.71} \\
	\end{tabular}
	\end{center}
	\caption{\label{font-table}OntoNotes NER}
	\label{table:ontonotes}
\end{table}

\begin{table*}[t!]
    \centering
    \small
	\begin{tabular}{lll}
	\textbf{the United Kingdom} & \textbf{Arab Foreign Ministers} & \textbf{Israeli Prime Minister Ehud Barak}\\
	\toprule
	the United States & Palestinian leaders & Italian President Francesco Cossiga\\
	the Tanzania United Republic & Yemeni authorities & French Foreign Minister Hubert Vedrine\\
	the Soviet Union &  Palestinian security officials & Palestinian leader Yasser Arafat\\
	the United Arab Emirates & Israeli officials & Iraqi leader Saddam Hussein\\
	the Hungary Republic & Canadian auto workers & Likud opposition leader Ariel Sharon\\
	Myanmar & Palestinian sources & UN Secretary General Kofi Annan\\
	Shanghai & many Jewish voters & Russian President Vladimir Putin\\
	China &  Lebanese Christian lawmakers & Syrian Foreign Minister Faruq al - Shara\\
	Syria & Israeli and Palestinian negotiators & PLO leader Arafat \\
	the Kyrgystan Republic & A Canadian bank & Libyan leader Muammar Gaddafi\\
	\end{tabular}
	\caption{Entity Embeddings Nearest Neighbours}
	\label{table:embeds}
\end{table*}

We also see a performance boost from using BERT embeddings, pushing the F1 up to 89.20. This falls slightly short of the state-of-the-art on this dataset, achieved using character-based Flair \cite{Flair} contextual embeddings. 

\section{Ablations}

To better understand the results, we conducted a small ablation study. 
The affect of including the \textbf{Static Layer} in the architecture is consistent across both datasets, yielding an improvement of around 2 F1 points; the updating of the token embeddings based on context seems to allow better merge decisions for each pair of tokens. Next, we look at the method used to update entity embeddings prior to combination into larger entities in the \textbf{Structure Layer}. In the described architecture, we use the Embed Update mechanism (see Figure~\ref{fig:levels_layer}), allowing embeddings to be changed dependent on which other embeddings they are about to be combined with. We see that this yields a significant improvement on both tasks of around 4 F1 points, relative to passing each embedding through a linear layer. 

The inclusion of an ``article theme'' embedding, used in the \textbf{Update Layer}, has little effect on the ACE05 data. but gives a notable improvement for OntoNotes. Given that the distribution of types of articles is similar for both datasets, we suggest this is due to the larger size of the OntoNotes set allowing the model to learn an informative article theme embedding without overfitting. 

\begin{table}[h]
\small
    \begin{center}
	\begin{tabular}{lcc}
		 & \textbf{ACE05} & \textbf{OntoNotes}\\[5pt]
		\midrule
		
		\textbf{Static Layer} & & \\
		\quad with & 74.6 &  87.59\\
		\quad without & 73.1 &  85.22\\
		
		\textbf{Embed Combination} & & \\
		\quad Linear & 70.2 & 83.96 \\
		\quad Embed Update & 74.6 & 87.59 \\
		
		\textbf{Article Embedding} & & \\
		\quad with & 74.5 & 87.59 \\
		\quad without & 74.6 & 85.60 \\
		
		\textbf{Sentence boundaries} & & \\
		\quad with & 70.8 &  86.30 \\
		\quad without & 74.6 &  87.59 \\
		
	\end{tabular}
	\end{center}
	\caption{\label{font-table}Architecture Ablations}
	\label{table:arch_ablations}
\end{table}

Next, we investigate the impact of allowing the model to attend to tokens in neighbouring sentences (we use a set kernel size of 30, allowing each token to consider up to 15 tokens prior and 15 after, regardless of sentence boundaries). Ignoring sentence boundaries boosts the results on ACE05 by around 4 F1 points, whilst having a smaller affect on OntoNotes. We hypothesize that this is due to the ACE05 task requiring the labelling of pronominal entities, such as ``he" and ``it", which is not required for OntoNotes. The coreference 
needed to correctly label their type is likely to require context beyond the sentence.

\section{Discussion}

\subsection{Entity Embeddings}

As our architecture merges multi-word entities, it not only outputs vectors of each word, but also for all entities - the tensor $T$. To demonstrate this, Table \ref{table:embeds} shows the ten closest entity vectors in the OntoNotes test data to the phrases ``the United Kingdom'', ``Arab Foreign Ministers" and ``Israeli Prime Minister Ehud Barak''.\footnote{Note that we exclude from the 10 nearest neighbours identical entities from higher levels. I.e.\ if ``the United Kingdom'' is kept as a three token entity, and not merged into a larger entity on higher levels, we do not report the same phrase from all levels in the nearest neighbours.}

Given that the OntoNotes NER task 
considers
countries and cities as GPE (Geo-Political Entities), the nearest neighbours in the left hand column are expected. The nearest neighbours of ``Arab Foreign Ministers'' and ``Israeli Prime Minister Ehud Barak'' are more interesting, as there is no label for groups of people or jobs for the task.\footnote{The phrase ``Israeli Prime Minister Ehud Barak" would have ``Israeli" labelled as NORP, and ``Ehud Barak" labelled as PERSON in the OntoNotes corpus.} Despite this, the model produces good embedding-based representations of these complex higher level entities. 


\subsection{Directional Embeddings} \label{dir_disc}

The representation of the \textbf{relationship between} each pair of words/entities as a vector is primarily a 
mechanism used by the model to update the word/entity vectors. However, the resulting vectors, corresponding to output \circled{D} of the Structure Layer, may also provide useful information for downstream tasks such as knowledge base population.

To demonstrate the directional embeddings, Table \ref{table:dir_embeds} shows the ten closest matches for the \textbf{direction} between ``the president" and ``the People's Bank of China". The network has clearly picked up on the relationship of an employee to an organisation. 

\begin{table}[h]
    \small
    \setlength{\tabcolsep}{0.6pt}
	\begin{tabular}{lcl}
		\textbf{the president} & $\rightarrow$ & \textbf{the People's Bank of China} \\
		\toprule
		the chairman & $\rightarrow$ & the SEC \\
		Vice Minister & $\rightarrow$ & the Ministry of Foreign Affairs \\
		Chairman & $\rightarrow$ & the People's Association of Taiwan \\
		Deputy Chairman & $\rightarrow$ & the TBAD Women's Division \\
		Chairman & $\rightarrow$ & the KMT \\
		Vice President & $\rightarrow$ & the Military Commission of the CCP \\
		vice-chairman & $\rightarrow$ & the CCP \\
		Associate Justices & $\rightarrow$ & the Supreme Court of the United States \\
		Chief Editor & $\rightarrow$ & Taiwan's contemporary monthly \\
		General Secretary & $\rightarrow$ & the Communist Party of China \\
	\end{tabular}
	\caption{Directional Embeddings Nearest Neighbours}
	\label{table:dir_embeds}
\end{table}

Table \ref{table:dir_embeds} also provides further examples of the network merging and providing intuitive embeddings for multi-word entities. 

\section{Conclusion}

We have presented a novel neural network architecture for smoothly merging token embeddings in a sentence into entity embeddings, across multiple levels. The architecture performs strongly on the task of nested NER, setting a new state-of-the-art F1 score by close to 8 F1 points, and is also competitive at flat NER. Despite being trained only for NER, the architecture provides intuitive embeddings for a variety of multi-word entities, a step which we suggest could prove useful for a variety of downstream tasks, including entity linking and coreference resolution. 

\begin{footnotesize}
\section*{Acknowledgments}
Andreas Vlachos is supported by the EPSRC grant eNeMILP
(EP/R021643/1).

\end{footnotesize}


\bibliography{acl2019}
\bibliographystyle{acl_natbib}

\clearpage

\appendix

\section{Supplemental Material}
\label{sec:supplemental}

\subsection{HyperParameters}

In addition to the hyperparameters recorded in the main paper, there are a large number of additional hyperparameters which we kept constant throughout experiments. The feedforward NN in the Static Layer, $FF_s$, has two hidden layers each of dimension 200. The NN in the Embed Update layer, $FF_{EU}$ has two hidden layers, each of dimension 320. The output NN has one hidden layer of dimension 200. Aside from $FF_{EU}$, which is initialized using the identity function as described in Supplementary section \ref{identity_init}, all parameters of networks are initialized from the uniform distribution between -0.1 and 0.1. The article theme size, a, is set to 50. All network layers use the SELU activation function of \cite{SELU}. The kernel size $k$ for the Static Layer is set to 6, allowing each token to attend the 3 tokens either side. 

On the OntoNotes Corpus, we train for 60 epochs, and half the learning rate every 12 epochs. On ACE 2005, we train for 150 epochs, and half the learning rate every 30 epochs. We train with a maximum batch dimension of 900 tokens. Articles longer than length 900 are split and processed in separate batches. We train using the Adam Optimizer, and, in addition to the dropout of 0.1, we apply a dropout to the Glove/LM embeddings of 0.2.

\subsection{Identity initialization} \label{identity_init}

\begin{figure}[h]
	\includegraphics[width=0.45\textwidth]{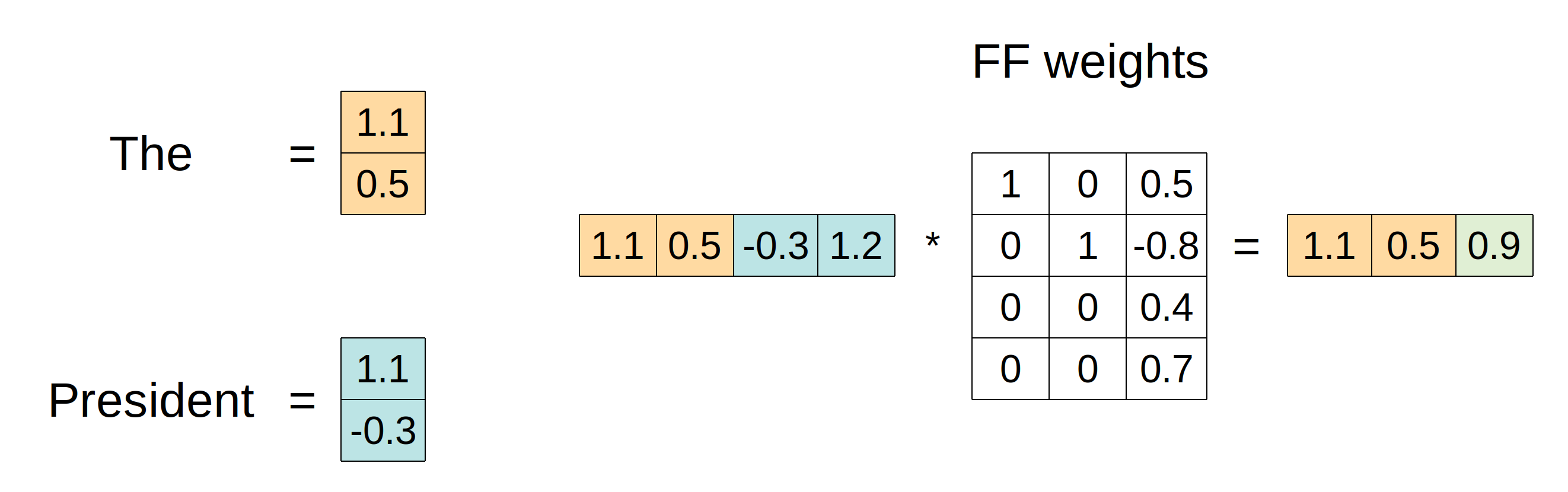}
	\caption{Update mechanism}
	\label{fig:identity}
\end{figure}

Figure \ref{fig:identity} gives a minimum working example of identity initialization of $FF_{EU}$. The embedding for ``The" is [1.1, 0.5], and that for ``President" is [1.1, -0.3]. Through the unfold ops, we'll end up with the two embeddings concatenated together. Figure \ref{fig:identity} shows $FF_{EU}$ as having just one layer with no activation function to demonstrate the effect of the identity initialization. The first two dimensions of the output are the embedding for ``The" with no changes. The final output (in light green) is the weighting.

In reality, the zeros in the weights tensor are initialized to very small random numbers (we use a uniform initialization between -0.01 and 0.01), so that during training $FF_{EU}$ learns to update the embedding for ``The" using the information that it is one step before the word ``President".

\subsection{Formation of outputs \circled{R} and \circled{D} in Structure Layer}

Outputs \circled{R} and \circled{D} of the \textbf{Structure Layer} have dimensions [b,s, k, e] and [b, s, k, d] respectively. These outputs are a weighted average of the directional and embedding outputs from the L levels of the structure layer. We use the weights, $W'$, (see Figure \ref{fig:sees}) to form the weighted average:

\[D = \sum_{l = 1}^{L} W'_l D_l\]

In the case of the weighted average for the embedding tensor, $R$, we use the weights from the next level. 

\[R = \sum_{l = 1}^{L} W'_{l+1} R_l\]

As a result, when updating, each token ``sees" information from tokens/entities on other levels dependent on whether or not they are in the same entity. For the intuition behind this, we use the example phrase ``The United Kingdom government" from Figure \ref{fig:sees}. The model should output merge values $M$ which group the tokens ``The United Kingdom" on the first level, and then group all the tokens on the second level. If this is the case, then for \textbf{the token ``United"}, $R$ and $D$ will hold the embedding of/directions to the tokens ``The" and ``Kingdom" in their disaggregated (unmerged) form. However, for \textbf{the token ``government"}, $R$ and $D$ will hold embeddings of/ directions to the combined entity ``the United Kingdom" in each of the three slots for ``The", ``United" and ``Kingdom". Because ``government" is not in the same entity as ``The United Kingdom" on the first level, it ``sees" the aggregated embedding of this entity.

Intuitively, this allows the token ``government" to update in the model based on the information that it has a country one step to the left of it, as opposed to having three separate tokens, one, two and three steps to the left respectively. Note that as with the entity merging, there are no hard decisions during training, with this effect based on the real valued merge tensor $M$, to allow differentiability.

\end{document}